\documentclass[letterpaper, 10 pt, conference]{ieeeconf}  

\IEEEoverridecommandlockouts                              

\usepackage[usenames,dvipsnames,table,xcdraw]{xcolor}  
\usepackage{hyperref}
\usepackage{amsmath,amsfonts}
\usepackage{algorithm}                    
\usepackage{algorithmicx}                  
\usepackage[noend]{algpseudocode}
\usepackage{graphicx}
\usepackage{multirow}
\usepackage{subcaption}

\newcommand{\dofs}{{\sc dof}s}
\newcommand{\cspace}{\ensuremath{\mathcal{C}_{space}}}
\newcommand{\cspaces}{\ensuremath{\mathcal{C}_{spaces}}}
\newcommand{\cfree}{\ensuremath{\mathcal{C}_{free}}}

\newcommand{\compcspace}{\ensuremath{\mathcal{C}_{comp}}}
\newcommand{\fk}{{\sc FK}}
\newcommand{\cc}{{\sc CC}}
\newcommand{\simd}{{\sc SIMD}}
\newcommand{\cpu}{{\sc CPU}}
\newcommand{\gpu}{{\sc GPU}}
\newcommand{\sbmp}{{\sc SBMP}}
\newcommand{\fc}{{\sc FFC}}
\newcommand{\mv}{{\sc MotVal}}
\newcommand{\fcl}{{\sc FCL}}
\newcommand{\cageenv}{{\sc Panda Cage}}

\newcommand{\crossenv}{{\sc Mobile Cross}}
\newcommand{\heteroenv}{\sc Heterogeneous}
\newcommand{\vamp}{{\sc VAMP}}
\newcommand{\vamrmp}{{\sc VA-MRMP}}
\newcommand{\seqsph}{{\sc SS}}
\newcommand{\ompl}{{\sc OMPL}}
\newcommand{\mrmp}{{\sc MRMP}}

\title{\LARGE \bf
Multi-Robot Motions in Milliseconds: Vector-Accelerated Primitives for Sampling-Based Planning
}

\author{
James D. Motes$^{1}$,  Marco Morales$^{1,2}$, and Nancy M. Amato$^{1}$
\thanks{$^{1}$James D. Motes, Marco Morales, and Nancy M. Amato are with the Parasol Lab, School of Computing and Data Science, University of Illinois at Urbana Champaign, Champaign, IL, 61820 USA.
{\tt\small jmotes2, moralesa, namato@illinois.edu}}%
\thanks{$^{2}$Marco Morales is also with the Department of Computer Science at Instituto Tecnol\'ogico Aut\'onomo de M\'exico (ITAM), Mexico City, M\'exico.}
}

\begin{document}

\maketitle
\thispagestyle{empty}
\pagestyle{empty}

\begin{abstract}

In this paper, we extend the recent Vector-Accelerated Motion Planning (\vamp) framework to multi-robot motion planning.
We develop two vector-accelerated primitives, multi-robot {\sc MotionValidation} ({\mv}) and {\sc FindFirstConflict} ({\fc}), which exploit {\simd} parallelism within the multi-robot domain.
On pure multi-robot motion validation tests, this achieves over \textcolor{black}{1415}$\times$ speedup in validation time.
Additionally, we modify a representative set of algorithms to use these new primitives.
\textcolor{black}{We evaluate five vector-accelerated multi-robot motion planning ({\vamrmp}) algorithms on manipulator, 2D mobile robot, and heterogeneous teams, observing planning speedups over {\fcl} of up to 1492$\times$.
With {\vamrmp}, all five planners attain subsecond median runtimes on problems with four Panda manipulators.}
\end{abstract}

\section{Introduction}
\label{sec:introduction}

In this paper, we explore efficient adaptations of the recent single-robot Vector-Accelerated Motion Planning ({\vamp})~\cite{thomason2024motions} techniques to the multi-robot domain. 
VAMP accelerates sampling-based motion planning ({\sbmp}) by exploiting single instruction/multiple data ({\simd}) forward kinematics ({\fk}) and collision checking ({\cc}) operations to validate multiple configurations in parallel on the {\cpu} and produces speedups of up to 500X on standard planning algorithms.
While much of this speedup comes from the parallelization (and the simplified sphere-based robot representation that facilitates it, a significant portion comes from intelligently exploiting parallel state validation to increase early termination during motion validation.
It is this key insight that we use to guide our design of vector-accelerated multi-robot motion planning ({\mrmp}).

We develop two {\mrmp} validation primitives: {\sc MotionValidation} ({\mv}), for boolean validation, and {\sc FindFirstConflict} ({\fc}), for discovering the earliest conflict.
Boolean validity is an extension of single robot validation with the additional concern of robot-robot collision at timesteps along synchronized motions in addition to standard robot-obstacle collisions.
A collision of any kind at any timestep invalidates the motion, and early termination can occur upon discovery of any collision.
Finding the first conflict is required by many algorithms which plan paths independently and then look for the earliest robot-robot collision to refine the paths~\cite{ssfs-cbsfomap-15,solis2024adaptive,sim2025st}.
As replanning paths can change the motions after a conflict, it is typically wasted effort to address conflicts occurring later in paths before resolving earlier conflicts which may remove or alter the later  conflicts.
As such, discovering a single collision witness does not allow for early termination, as all timesteps before this collision must be checked to ensure it is the earliest conflict.

\begin{table}[t]
\caption{Algorithm Taxonomy and Four-Panda Results}
\setlength{\tabcolsep}{3.1pt}
\centering
\begin{tabular}{|l|c|c|c|c|c|}
\hline
\multicolumn{1}{|c|}{Algorithm}
& \begin{tabular}[c]{@{}c@{}}Composite\\RRT-C\end{tabular}
& \begin{tabular}[c]{@{}c@{}}PP-\\ST-RRT\end{tabular}
& \begin{tabular}[c]{@{}c@{}}MR-\\dRRT\end{tabular}
& ST-CBS
& ARC \\ \hline

Representation
& C & D & D & D & H \\ \hline

Search
& C & D & C & H & H \\ \hline

Validity Query
& MV & MV & MV & FFC & MV/FFC \\ \hline

Runtime (ms)
& 152
& 189
& 813
& 247
& 50
\\ \hline

Speedup
& $1492\times$
& $550\times$
& $138\times$
& $213\times$
& $626\times$
\\ \hline
\end{tabular}

\caption*{\small
Representation/Search: Coupled (C), Decoupled (D), Hybrid (H)\\
Validity Query: {\sc MotionValidation} (MV),
{\sc FindFirstConflict} (FFC)\\
\textcolor{black}{Four-Panda results use the {\sc Panda Cage} benchmark with
$N=50$ trials.
Runtime is the median solve timeover
successful {\vamrmp} trials.
Speedups are paired geometric means of per-seed {\fcl}/{\vamrmp} runtime ratios over trials solved by both implementations.
See Sec.~\ref{sec:evaluation} for complete results.}}

\label{tab:algs}
\end{table}

\begin{figure}[t]
\vspace{-1em}
    \centering
    \begin{subfigure}[]{0.235\textwidth}
        \centering
        \includegraphics[width=\linewidth]{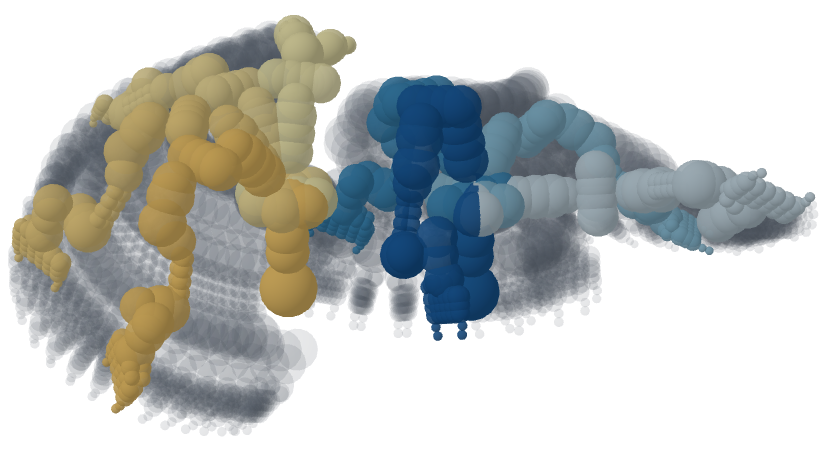}
        \caption{Rake - No collision}
        \label{fig:rake-free}
    \end{subfigure}
    \hfill
    \begin{subfigure}[]{0.235\textwidth}
        \centering
        \includegraphics[width=\linewidth]{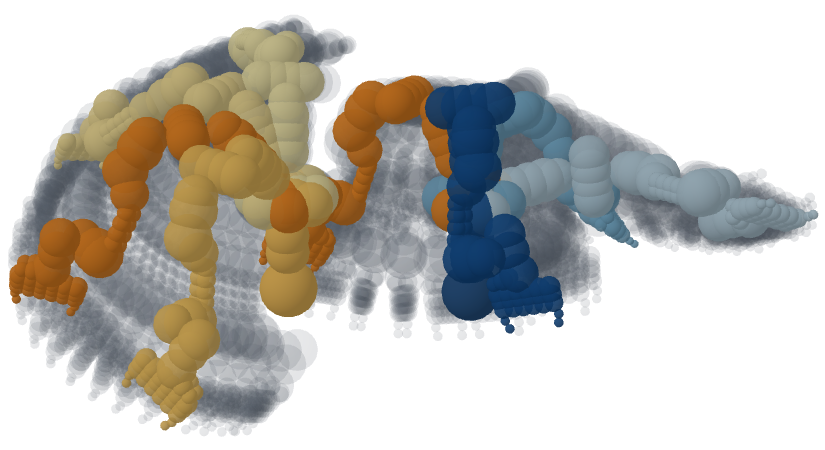}
        \caption{Rake - Collision}
        \label{fig:rake-collision}
    \end{subfigure}
    \caption{
    \small
    The multi-robot {\mv} primitive utilizes the \textit{rake} technique to check for robot-robot collisions in parallel between configurations at timesteps distributed in time over the motions.
    The paths of the blue and yellow robots are shown with the gradient indicating later timesteps in darker colors.
    Fig.~\ref{fig:rake-free} shows the initial rake simultaneously \textcolor{black}{checking timesteps in the batch} $\{0,\frac{t_{max}}{4},\frac{2t_{max}}{4},\frac{3t_{max}}{4}$\} for robot-robot collision.
    Fig.~\ref{fig:rake-collision} shows a conflict in orange discovered quickly on the third step of the rake, at a timestep in the second half of the motion.
    \textcolor{black}{Note, a four-lane rake is shown for clarity; the implementation uses $W=8$.}
    }
    \vspace{-2em}
    \label{fig:rake}
\end{figure}

For both primitives, we develop strategies to improve early termination while exploiting {\simd} parallelism for {\fk} and {\cc} operations. 
We integrate these primitives into a set of {\mrmp} methods representative of the different uses of validation and evaluate the impact on manipulator, mobile robot, and heterogeneous team scenarios.
\textcolor{black}{
The primitives reduce pure motion validation time by up to 1415$\times$ and planning time over {\fcl} by up to 1492$\times$ on matched successful trials.
}
The included methods (and selected results) are listed in Table~\ref{tab:algs}.
In summary, our contributions are:
\begin{itemize}
    \item Vector-accelerated multi-robot motion planning ({\vamrmp}) primitives: {\mv} and {\fc}
    \item Integration of these primitives into representative sampling-based {\mrmp} algorithms
    \item Empirical evaluation across manipulator, mobile robot, and heterogeneous robot teams
    \item Open-source release of {\vamrmp} implementations (upon acceptance)
\end{itemize}

\section{Background and Related Work}
\label{sec:related-work}

In this section, we define the motion planning and multi-robot motion planning ({\mrmp}) problems and discuss algorithms and vector-acceleration for both. 

\subsection{Motion Planning}
\label{sec:motion-planning}

A robot configuration is defined by its degrees-of-freedom ({\dofs}).
The configuration space {\cspace} is the set of all robot configurations in an environment~\cite{lw-apcfpapo-79}.
Motion planning seeks a continuous path through the free space {\cfree}.
Representing {\cspace} explicitly is intractable in the general case~\cite{ss-otpmpiigtfctporam-83,c-crmp-88}.
Sampling-based motion planning ({\sbmp}) instead samples random configurations and motions between them to search for a path~\cite{kslo-prpp-96,kuffner2000rrt}.
Configurations are validated by performing forward kinematics ({\fk}) and collision checking ({\cc}), and motions are validated by discretizing them into intermediate configurations which are then validated.

\textcolor{black}{
Within {\sbmp}, for both single start/goal query algorithms~\cite{kuffner2000rrt} and multi-query algorithms~\cite{kslo-prpp-96}, the motion validation process has historically been the bottleneck.
Methods like the recursive midpoint bisection checking~\cite{sanchez2003single} propose strategies on \textit{where to check} which aim to quickly find a witness collision and improve early termination.
Others choose \textit{when to check}, often deferring validation until a motion is known to be required~\cite{bk-ppulp-00}.
Finally, some methods improve \textit{how many checks can be done at once}~\cite{thomason2024motions}.
We pull insights from all three for the multi-robot domain.
}

\subsection{Vector-Accelerated Motion Planning}

Vector-Accelerated Motion Planning (VAMP)~\cite{thomason2024motions} introduced fine-grained parallelism to {\sbmp} by exploiting {\simd} operations in primitive functions common to {\sbmp} algorithms.
Much of the {\simd} modification focused on {\fk} and {\cc} where batches of configurations can be processed in parallel on the {\cpu}, enabling high throughput with low latency.
These {\simd} {\fk} and {\cc} operations are enabled by adopting a sphere-based representation of a robot.
This sphere-based representation is then given to a tracing compiler which builds specialized {\simd} kernels for that specific robot's {\fk} and {\cc} operations.
In addition to enabling the {\simd} operations, this simplified geometric representation contributes to the overall speedup demonstrated in~\cite{thomason2024motions}.

\textcolor{black}{
Beyond the speedup from processing batches of configurations simultaneously, {\vamp} proposes intelligent construction of these batches to increase early termination during motion validation.
Instead of linearly scanning from one end of a motion to the other in batches of consecutive configurations, batches are created from configurations \textit{spatially distributed} evenly along the motion.
The validation of these distributed batches is referred to as a \textit{rake}.
This is an application of the insights from prior sequential validation methods that collisions can be discovered faster by distributing {\cc} along the motion to the batch processing offered by {\simd} operations.
More details are provided in the multi-robot application of the rake approach in Sec.~\ref{sec:cc-as-validity} and Fig.~\ref{fig:rake}.
}

\textcolor{black}{
{\gpu} planners such as cuRobo~\cite{sundaralingam2023curobo} and pRRTC~\cite{huang2025prrtc} exploit SIMT parallelism over large batches of optimization seeds, tree expansions, and {\cc}s, providing high throughput when sufficient work remains on-device. 
Like {\vamp}, our {\cpu} {\simd} primitives target a complementary regime of low-latency, irregular queries: they run on standard {\cpu}s without a discrete accelerator, incur no {\gpu} kernel-launch or host-device communication overhead, and support early termination between small timestep and robot-pair batches.
}

\subsection{Multi-robot Motion Planning}
\label{sec:mrmp}

Multi-robot motion planning ({\mrmp}) seeks to find a collision-free path through the composite configuration space $\compcspace$ which is the Cartesian product of the individual robot $\cspaces$.
Most {\mrmp} papers focus on addressing the exponential growth of $\compcspace$ with the number of robots.
Here, instead we focus on the cost of validation.

A composite configuration is invalid if any individual robot configuration is in collision with an obstacle, itself, or another robot.
This cost quickly gets dominated by the robot-robot {\cc}s which scale quadratically with the number of robots while robot-obstacle {\cc}s scale linearly. 
Multi-robot motion validation is still computed by discretizing individual motions into intermediate configurations. 
However, each discrete intermediate corresponds to one timestep as robot-robot collisions occur in both space and time.

There are two primary multi-robot validity queries: boolean validity and find the first conflict.
The boolean validity query simply returns a boolean true/false validity for the entirety of the set of synchronized robot motions.
This is primarily used by methods which directly apply single robot {\sbmp} to the multi-robot domain either by directly searching $\compcspace$~\cite{sl-uppccdpmrs-2002,ssh-faniaehdrfeoirimm-16}, or treating other robots as dynamic obstacles~\cite{kerimov2025si}.
These motions can be invalidated by either robot-obstacle or robot-robot collisions.
Some methods consider these simultaneously~\cite{sl-uppccdpmrs-2002},
while others first compute decoupled representations valid in the individual robot $\cspaces$ and constrain multi-robot motions to compositions of these~\cite{ssh-faniaehdrfeoirimm-16}, decoupling robot-obstacle and robot-robot validation (deciding \textit{when to validate}).

Methods which query for the first conflict typically compute entire paths valid in the individual robot $\cspaces$ and then use robot-robot collisions (and the timesteps they occur at) to guide either search~\cite{smsa-rmmpucs-21,sim2025st} or representation construction~\cite{solis2024adaptive,sim2025st}.
This validity query assumes individual validity and only checks for synchronized robot-robot collisions, so no robot-obstacle {\cc}s are needed.
It requires the \textit{first} conflict in the set of paths as conflict resolution often de-synchronizes later portions of the paths, undoing prior resolution effort.
Thus, a single robot-robot witness is insufficient, and all timesteps prior to a conflict must be validated before a conflict can be determined to be the earliest.
It is from the difference between these two validity queries that we claim the need for two distinct multi-robot validity primitives: {\mv} and {\fc} based on their query and use of validity information, and categorize {\mrmp} algorithms by which of these they use (Table~\ref{tab:algs}).

We find that this categorization naturally extends the typical taxonomy of {\mrmp} methods into coupled, decoupled, or hybrid approaches to either representation construction or search (Table~\ref{tab:algs}).
In fact, the validity primitive is central to the implementation of these other design choices.
For example, all coupled searches must use {\mv} to validate any motion segment in $\compcspace$~\cite{sl-uppccdpmrs-2002} (even when individual validity is computed separately in decoupled representations~\cite{ssh-faniaehdrfeoirimm-16}).
Even decoupled searches like prioritized planning still use a variant of {\mv} to ensure their individual paths do not collide with higher priority robot motions~\cite{kerimov2025si}.
Meanwhile, hybrid methods which typically start decoupled and then \textit{decide} when to couple often use conflict information to guide these decisions~\cite{solis2024adaptive,sim2025st}. 
When the representations stay decoupled~\cite{smsa-rmmpucs-21,sim2025st}, only {\fc} is needed, but when hybrid approaches build representations in different compositions of individual robot $\cspaces$, {\mv} is again needed to validate the motion segments in the representation~\cite{solis2024adaptive}.

The set of methods in Table~\ref{tab:algs} provides a representative for each combination of these design choices with varying strategies for computing robot-obstacle and robot-robot validity.  
In Sec.~\ref{sec:applications} we describe the application of the new {\mv} and {\fc} (and existing {\vamp}) primitives to these methods and evaluate the impact in Sec.~\ref{sec:evaluation}.
Many of these methods have extensions and improvements, but we attempted to choose a method representing the root of a method family.

\subsection{Vector-Accelerated Multi-Robot Motion Planning}
\label{sec:vec-mrmp}

\textcolor{black}{
Concurrent work~\cite{huang2026vampmr} adapts {\vamp}~\cite{thomason2024motions} to multi-manipulator planning and applies it to Composite RRT-C and CBS-MP~\cite{smsa-rmmpucs-21}. 
In contrast, our contribution is the explicit separation of boolean {\mv} from the earliest conflict {\fc} based on their different information requirements and the corresponding change in batch packing behavior.
We further compare combined and hierarchical ordering of robot-obstacle and robot-robot checks across a motion and evaluate both primitives across five coupled, decoupled, and hybrid planners on manipulator, rigid-body, and heterogeneous teams with increasing robot counts. 
}

\section{Methodology}
\label{sec:method}
In this section, we provide two vector-accelerated {\mrmp} primitives: {\mv} (Sec.~\ref{sec:cc-as-validity}) and {\fc} (Sec.~\ref{sec:cc-as-info}).
We then describe how a representative set of {\mrmp} algorithms can utilize these modified primitives (Sec.~\ref{sec:applications}).

\subsection{Motion Validation}
\label{sec:cc-as-validity}

The multi-robot {\mv} primitive (Alg.~\ref{alg:mv}) is used when the algorithm cares only for the boolean validity of a set of robot paths $P$.
As described in Section~\ref{sec:mrmp}, this is usually in coupled approaches where the movement of all robots is being considered at once.
A variant of it is also used in decoupled, prioritized planning when the validity of motion segments for a single robot needs to be checked against the already planned motions of higher priority robots (discussed in Section~\ref{sec:prioritized}).
\textcolor{black}{In either case, the objective is to discover a witness collision as early as possible if one exists, optimizing early termination.
This can be achieved by either a robot-obstacle collision or a robot-robot collision.
}

\textcolor{black}{
During empirical evaluation, we have seen the likelihood of which collision type can be discovered faster vary with the number of robots and the density of obstacles, so we present two different strategies: \textit{combined} and \textit{hierarchical}.
The combined approach checks robot-obstacle and robot-robot validity for each batch of timesteps before moving onto the next batch of timesteps
(this is a straightforward application of the state-based validation discussed in~\cite{huang2026vampmr}).
Hierarchical first performs all robot-obstacle validation and then all robot-robot validation.
Intuitively, robot-obstacle collisions are more likely when there are more obstacles or more robots to collide with those obstacles.
Note that robot-obstacle validation also checks the robot for self-collision.
We provide a study of the different strategies in Sec.~\ref{sec:validation-timing}.
}

\textcolor{black}{
Alg.~\ref{alg:mv} uses a set of underlying {\vamp} functions.
{\sc PackCfgBatch} takes in a path $p_i$, a batch index $b$, and a \textit{packing strategy} (either \textit{rake} or \textit{linear}).
The rake strategy temporally distributes timesteps evenly along the path (as is done with the spatial distribution in {\vamp}~\cite{thomason2024motions}).
For batch index $b$, \textcolor{black}{{\simd} vector-width of $W$}, and \textcolor{black}{$\textit{num\_batches}=\lceil\frac{\textit{total timesteps}}{W}\rceil$, this would result in $B_i$ consisting of the configurations along $p_i$ at timesteps $b, b+\textit{num\_batches}, b+\textit{num\_batches}*2,...,b+\textit{num\_batches}*(W-1)$.}
The linear strategy results in the batch packed with configurations at consecutive timesteps in $p_i$, \textcolor{black}{$B_i=\{b*W, b*W+1,... b*W+W-1\}$}.
See Fig.~\ref{fig:rake} for an example of the rake strategy.
\textcolor{black}{Note, paths with fewer timesteps are padded with their last configuration to prevent the disappearing robot problem.}
}

Regardless of packing strategy, {\sc SpheresFK} takes batches of configurations and uses the robot-specific {\fk} kernel to return the positions of each sphere $S_i$ (from the sphere-based robot representation) for every configuration in $B_i$.
The {\cc} function then checks each set of spheres in $S_i^\textit{local}$ for collision with \textcolor{black}{the set of simple geometries representing the environment obstacles} (or itself), returning true if any configuration in the batch is in collision.
\textcolor{black}{Note, {\cc} is again used to denote the collision checking of two sets of simple geometries $S_i^\textit{world}$, $S_j^\textit{world}$ (spheres) for robot-robot collision.}

\textcolor{black}{
Alg.~\ref{alg:mv} begins with the switch on hierarchical behavior on line~\ref{alg:mv:if-hierarchical} after computing the total number of batches needed to cover the set of paths (line~\ref{alg:mv:num-batches}).
In the hierarchical strategy, the validation rakes through all paths simultaneously, checking validity of each robot for the configurations at the timesteps in batch $B_i$ before incrementing the batch (lines~\ref{alg:mv:env-check-start}-\ref{alg:mv:env-check-end}).
Any discovered collision results in early termination (lines~\ref{alg:mv:cc-env}, \ref{alg:mv:env-termination}).
}

\textcolor{black}{
Both strategies then iterate over all batches (line~\ref{alg:mv:rob-batch-loop}).
The combined strategy first validates all individual configurations (lines~\ref{alg:mv:env-check-start-2}-\ref{alg:mv:env-check-end-2}), terminating early upon any discovered collision.
Both strategies then check for robot-robot collision between all pairs $r_i,r_j$ for the temporally distributed timesteps of batch $b$ packed using the rake strategy (lines~\ref{alg:mv:rake-rob-1}, \ref{alg:mv:rake-rob-2}), terminating early if any collision is found (line~\ref{alg:mv:rob-termination}).
If all validation checks pass, the algorithm returns true (line~\ref{alg:mv:valid}).
}

\textcolor{black}{
There is an additional design choice of iterating over entire paths or pairs of paths for individual or robot-robot validation instead of raking across all paths at once.
Empirically, we found that the simultaneous rake makes a large difference for the individual validation and a smaller, but still meaningful difference for the robot-robot validation.
Future work may examine mixed strategies which identify pairs of robots that are likely to collide (e.g., adjacent fixed-base manipulators), and prioritize those pairs before broader validation. 
}

Note, for fixed-base robots, the {\sc SpheresFK} calls on lines~\ref{alg:mv:spheres-fk-env} and~\ref{alg:mv:spheres-fk-env-2} return the robot's representation spheres in the robot reference frame and the environment obstacles are transformed into this frame (in practice, the obstacle transformations are computed once and cached for each robot).
The {\sc SpheresFK} calls on lines~\ref{alg:mv:spheres-fk-rob-1}, \ref{alg:mv:spheres-fk-rob-2} return the robots' representation spheres in the world frame, so they can be checked against each other.
This approach is also discussed in~\cite{huang2026vampmr}.

\begin{algorithm}[t]
\footnotesize
\caption{\textsc{MotionValidation} ({\mv})}
\label{alg:mv}
\begin{algorithmic}[1]
    \Statex \textbf{Input:} Paths $P=\{p_1,\dots,p_n\}$, \textcolor{black}{environment obstacle geometries per robot local frame $E=\{E_1,\dots, E_n\}$}, \textit{hierarchical}, \textit{\textit{check\_environment}}
    \State $t_{\max} \gets \max_{p_i \in P} |p_i|$
    \State \textcolor{black}{$\textit{num\_batches} \gets \left\lceil \dfrac{t_{\max}}{W} \right\rceil$}\label{alg:mv:num-batches}
    \If{\textit{hierarchical} and \textit{check\_environment}}\label{alg:mv:if-hierarchical}
        \For{$b \in \{0,\dots,\textit{num\_batches}-1\}$}\label{alg:mv:env-check-start}
            \For{$i \in \{1,\dots,n\}$}
                \State Batch $B_i \gets \Call{PackCfgBatch}{p_i, b,\textit{rake}}$\label{alg:mv:rake-env}
                \State \textcolor{black}{Spheres $S_i^\textit{local} \gets \Call{SpheresFK}{r_i, B_i}$}\label{alg:mv:spheres-fk-env}
                \If{\textcolor{black}{$\textit{collision} \gets \Call{CC}{S_i^\textit{local}, E_i}$}}\label{alg:mv:cc-env}
                    \State \Return \textbf{false}\label{alg:mv:env-termination}
                \EndIf
            \EndFor
        \EndFor\label{alg:mv:env-check-end}
    \EndIf
    \For{$b \in \{0,\dots,\textit{num\_batches}-1\}$}\label{alg:mv:rob-batch-loop}
        \If{not \textit{hierarchical} and \textit{check\_environment}}
            \For{$i \in \{1,\dots,n\}$}\label{alg:mv:env-check-start-2}
                \State Batch $B_i \gets \Call{PackCfgBatch}{p_i, b,\textit{rake}}$\label{alg:mv:rake-env-2}
                    \State \textcolor{black}{Spheres $S_i^\textit{local} \gets \Call{SpheresFK}{r_i, B_i}$}\label{alg:mv:spheres-fk-env-2}
                    \If{\textcolor{black}{$\textit{collision} \gets \Call{CC}{S_i^\textit{local}, E_i}$}}\label{alg:mv:cc-env-2}
                        \State \Return \textbf{false}
                    \EndIf
            \EndFor\label{alg:mv:env-check-end-2}
        \EndIf
        \For{$i \in \{1,\dots,n-1\}$}\label{alg:mv:rob-check-start}\label{alg:mv:outer-loop}
            \State Batch $B_i \gets \Call{PackCfgBatch}{p_i, b,\textit{rake}}$\label{alg:mv:rake-rob-1}
            \State \textcolor{black}{Spheres $S_i^\textit{world} \gets \Call{SpheresFK}{r_i, B_i}$}\label{alg:mv:spheres-fk-rob-1}
            \For{$j \in \{i+1,\dots,n\}$}\label{alg:mv:inner-loop}
                \State Batch $B_j \gets \Call{PackCfgBatch}{p_j, b,\textit{rake}}$\label{alg:mv:rake-rob-2}
                \State \textcolor{black}{Spheres $S_j^\textit{world} \gets \Call{SpheresFK}{r_j, B_j}$}\label{alg:mv:spheres-fk-rob-2}
                \If{\textcolor{black}{$\textit{collision} \gets \Call{CC}{S_i^\textit{world}, S_j^\textit{world}}$}}\label{alg:mv:cc-rob}
                    \State \Return \textbf{false}\label{alg:mv:rob-termination}
                \EndIf
            \EndFor
        \EndFor
    \EndFor\label{alg:mv:rob-check-end}
    \State \Return \textbf{true}\label{alg:mv:valid}
\end{algorithmic}
\end{algorithm}

\subsection{Find First Conflict}
\label{sec:cc-as-info}

\begin{algorithm}
\footnotesize
\caption{\textsc{FindFirstConflict} ({\fc})}
\label{alg:fc}
\begin{algorithmic}[1]
    \Statex \textbf{Input:} Paths $P=\{p_1,\dots,p_n\}$
    
    \State $t_{\max} \gets \max_{p_i \in P} |p_i|$
    \State \textcolor{black}{$\textit{num\_batches} \gets \left\lceil \dfrac{t_{\max}}{W} \right\rceil$}\Comment{\textcolor{black}{$W$ is the {\simd} vector-width}}
    \State Conflict $C\gets\textbf{null}$

    \For{$b\in \{0,\dots,\textit{num\_batches}-1\}$}
        \State \textcolor{black}{$t_{\textit{batch\_start}} \gets b \cdot W$}
        \For{$i\in\{1,\dots, n-1\}$}
            \State \textcolor{black}{Batch $B_i \gets \Call{PackCfgBatch}{p_i, b,\textit{linear}}$}\label{alg:fc:rake-rob-1}
            \State \textcolor{black}{Spheres $S_i^\textit{world} \gets \Call{SpheresFK}{r_i, B_i}$}\label{alg:fk:spheres-fk-rob-1}
            \For{$j\in\{i+1,\dots, n\}$}
                \State \textcolor{black}{Batch $B_j \gets \Call{PackCfgBatch}{p_j, b,\textit{linear}}$}\label{alg:fc:rake-rob-2}
                \State \textcolor{black}{Spheres $S_j^\textit{world} \gets \Call{SpheresFK}{r_j, B_j}$}\label{alg:fk:spheres-fk-rob-2}
                \State \textcolor{black}{$t_\textit{conflict} \gets \Call{HasConflict}{S_i^\textit{world}, S_j^\textit{world}, t_{\textit{batch\_start}}}$}
                \Comment{returns first collision timestep in the batch, offset by $t_{\text{batch\_start}}$}
                \If{\textcolor{black}{$t_\textit{conflict}=t_\textit{batch\_start}$}}\Comment{\textcolor{black}{Must be earliest}}\label{alg:fc:conflict-at-first-timestep}
                    \State\textcolor{black}{\Return $C\gets\{t_{\textit{conflict}}, r_i, r_j\}$}
                \EndIf
                \If{$t_\textit{conflict} \neq \textbf{null}$} 
                    \If {$C = \textbf{null}$ or $C.t > t_{\textit{conflict}}$}
                        \State $C\gets\{t_{\textit{conflict}}, r_i, r_j\}$
                    \EndIf
                \EndIf
            \EndFor
        \EndFor
        \If{$C\neq\textbf{null}$}
        \State\Return $C$
        \EndIf
    \EndFor
    \State \Return \textbf{null}
\end{algorithmic}
\end{algorithm}

The {\fc} primitive is used when collision information is required to guide either representation construction or search (typically a hybrid method of some kind).
This requires discovering the \textit{first} timestep at which any robot-robot collision occurs in the set of paths $P$ \textcolor{black}{(rather than any collision witness)}.

In addition to the return type, there are two main distinctions between {\mv} in Alg.~\ref{alg:mv} and {\fc} in Alg.~\ref{alg:fc}.
First, {\fc} does not consider robot-obstacle collisions.
It assumes that all paths are free of robot-obstacle collisions and is only searching for robot-robot collisions (all {\sc SpheresFK} calls return spheres in the world frame).

\textcolor{black}{
Second, it uses the \textit{linear} strategy for packing configuration batches (lines~\ref{alg:fc:rake-rob-1}, ~\ref{alg:fc:rake-rob-2}).
As a conflict at timestep $t_\textit{conflict}$ can only be confirmed as the first conflict if all timesteps $t<t_\textit{conflict}$ are collision-free, validation must be performed for timesteps $0,...,t_\textit{conflict}$.
Thus, employing the rake strategy which temporally distributes timestep batches evenly along the paths will result in validation effort spent on timesteps $t>t_\textit{conflict}$ (with the exception of $t_\textit{conflict}=t_\textit{max}$).
Taking a linear packing approach ensures the minimum number of timestep batches are validated before discovering $t_\textit{conflict}$.
For simplicity of notation, we leave the extraction of the individual timestep $t_\textit{conflict}$ within the batch of consecutive configurations in the \textcolor{black}{{\sc HasConflict}} function call.
}

\subsection{Application in MRMP Algorithms}
\label{sec:applications}
Here we provide a brief description of how the representative set of algorithms in Table~\ref{tab:algs} use our new multi-robot primitives or primitives from single robot {\vamp}~\cite{thomason2024motions}.
The ``Validity Query'' row in Table~\ref{tab:algs} provides a quick summary for which algorithms use either {\mv} or {\fc}.

\subsubsection{Composite RRT-C}
\label{sec:composite}

Composite RRT-C is a direct application of the single robot RRT-C~\cite{kuffner2000rrt} in {\compcspace}.
The implementation mirrors the RRT-C implementation in~\cite{thomason2024motions} with the multi-robot {\mv} primitive replacing the single robot version.

\subsubsection{PP-ST-RRT}
\label{sec:prioritized}

Prioritized planning with Space Time RRT~\cite{grothe2022st} plans for robots sequentially in a fixed priority order (referred to as PP-ST-RRT)~\cite{kerimov2025si}.
Higher priority paths are treated as dynamic obstacles for lower priority paths using ST-RRT.
The {\mv} primitive in ST-RRT is replaced with a modified version of Alg.~\ref{alg:mv}.
As PP-ST-RRT plans for one robot at a time, the outer loop on line~\ref{alg:mv:outer-loop} is replaced with only the current robot and the inner loop on line~\ref{alg:mv:inner-loop} iterates over only higher priority robots.
We simplify the implementation of ST-RRT here and do not include the rewiring mechanisms as many of the other algorithms seek only to find a valid path, though we expect the relative performance gains to hold.

\subsubsection{ST-CBS}
\label{sec:st-cbs}

ST-CBS~\cite{sim2025st} modifies the ST-RRT~\cite{grothe2022st} logic to work within the Conflict-Based Search (CBS) framework~\cite{ssfs-cbsfomap-15}.
CBS builds a \textit{constraint tree} (CT) where nodes contain sets of constraints for each robot and paths which adhere to those constraints.
The search evaluates the paths in a node for conflicts, and, if none are discovered, returns a valid solution.
Upon discovering a conflict between robots $r_i,r_j$, a constraint is generated for each, and the CT branches, creating two new nodes, one with an additional constraint for $r_i$ and one with an additional constraint for $r_j$.

ST-CBS utilizes a unidirectional ST-RRT (UST-RRT) method to query individual paths.
The basic primitives function the same as those in ST-RRT with constraints forming obstacles in space-time which can be represented as instantaneous (or short-lived) dynamic obstacles.
Additionally, the new {\fc} primitive is employed to search for conflicts between paths when evaluating nodes in the CT.

\subsubsection{MR-dRRT}
\label{sec:dRRT}

MR-dRRT~\cite{ssh-faniaehdrfeoirimm-16} begins by building roadmaps for each individual robot.
All relevant primitives here are replaced by VAMP~\cite{thomason2024motions}.
The discrete RRT search over the tensor product graph of the roadmaps uses the multi-robot {\mv} primitive to validate composite edges (skipping robot-obstacle {\cc}).
Finally, the connect-to-target function (extend a composite state directly to the goal) uses the {\mv} with robot-obstacle {\cc}.

\subsubsection{ARC}
\label{sec:arc}

ARC~\cite{solis2024adaptive} first builds a path for each individual robot.
The original paper uses PRM, but we replace this with a VAMP RRT-C.
The {\fc} primitive is then repeatedly called on these paths.
Upon discovering a conflict, local subproblems are introduced to replan the conflicting paths in the conflicting region.
The original paper proposes a configurable hierarchy of methods for solving the subproblems.
We use a single instance of Composite RRT-C.

\section{Evaluation}
\label{sec:evaluation}

We conduct a set of experiments to evaluate the impact of the vector-accelerated multi-robot primitives on both pure validation and {\mrmp} algorithm performance.
Section ~\ref{sec:experimental-design} describes our experimental design and Sections~\ref{sec:validation-timing} and~\ref{sec:planner-exp} report the results of each experiment.

\begin{figure*}[t]
    \centering
        \hfill
    \begin{subfigure}[t]{0.30\textwidth}
        \centering
        \includegraphics[width=.5\linewidth]{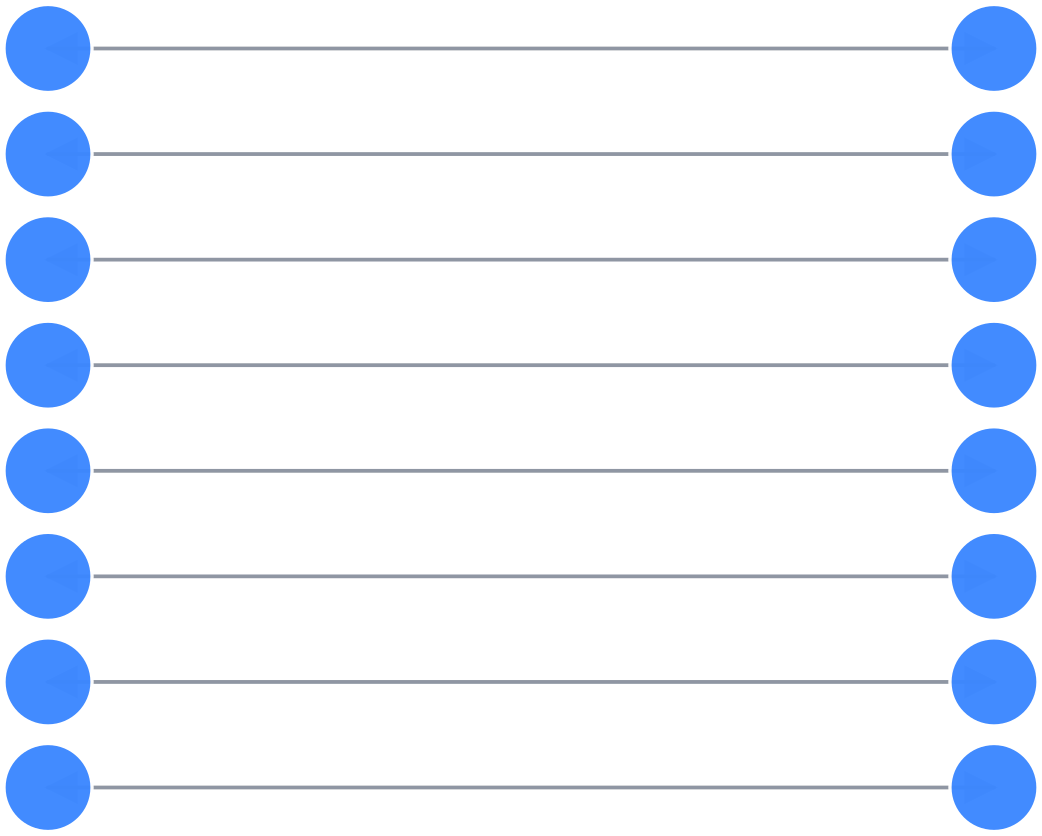}
        \caption{{\crossenv} Scenario}
        \label{fig:cage}
    \end{subfigure}
    \hfill
    \begin{subfigure}[t]{0.3\textwidth}
        \centering
        \includegraphics[width=0.5\linewidth]{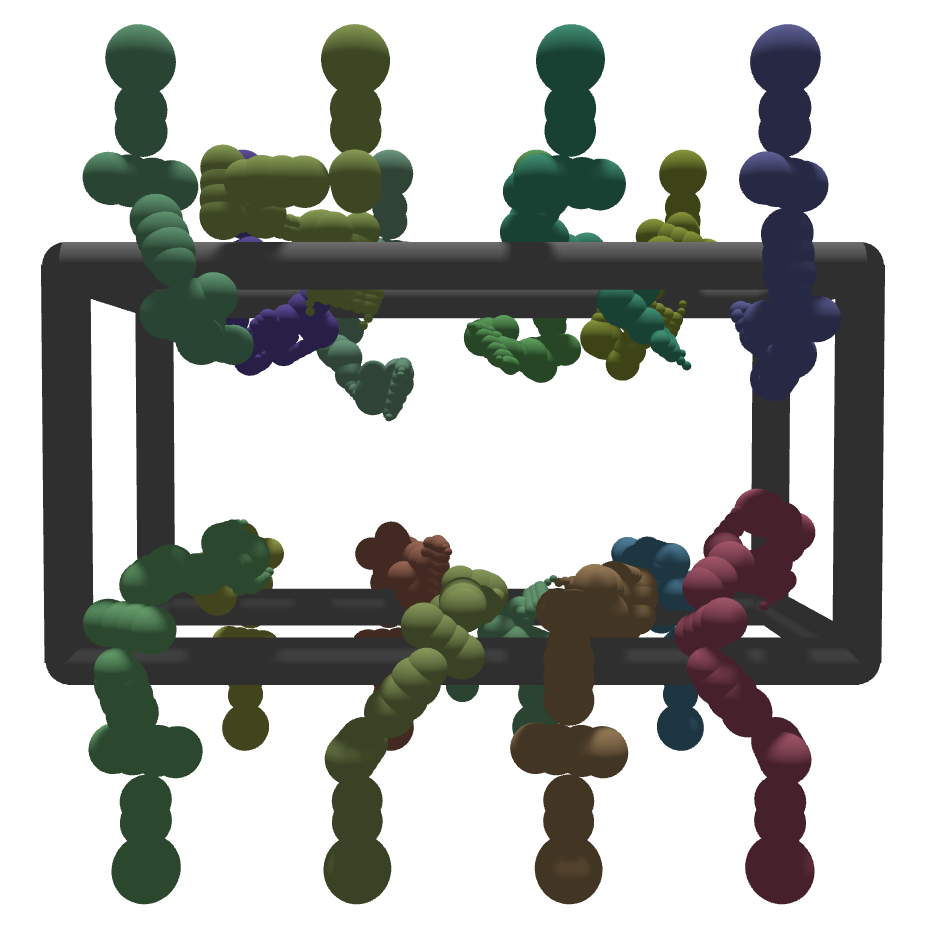}
        \caption{{\cageenv} Scenario}
        \label{fig:cage}
    \end{subfigure}    \hfill
    \begin{subfigure}[t]{0.3\textwidth}
        \centering
        \includegraphics[width=1.0\linewidth]{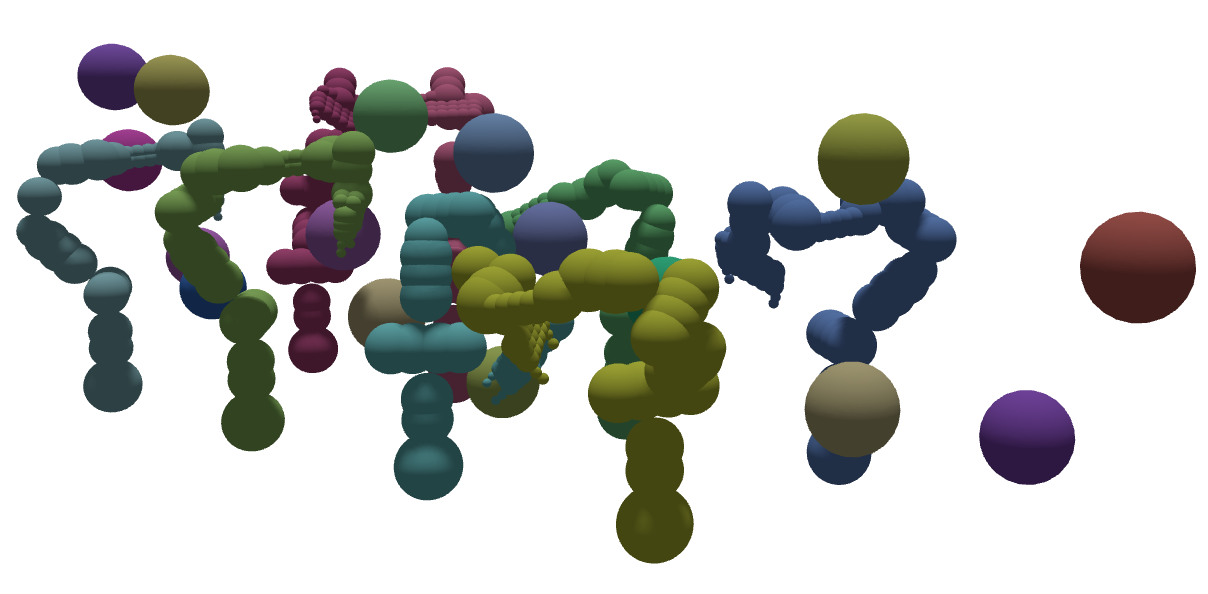}
        \caption{{\heteroenv} Scenario}
        \label{fig:hetero}
    \end{subfigure}
    \hfill
    \caption{
    \small
    \textcolor{black}{(a) Two rows of mobile robots swap places in an open environment forcing n/2 conflicts.}
    (b) {\cageenv} environment with 16 manipulators operating in a shared workspace arranged in floor- and ceiling-mounting (e.g., automotive manufacturing).
    A random start and goal position is sampled for each manipulator with the end effector inside the cage.
    (c) 8 Panda manipulators form a corridor. A random start is sampled for each manipulator, and a goal pose is selected with the end-effector inside the corridor.
    16 flying sphere robots, split into two groups, must swap places on either side of the corridor.
    The corridor is limited to the reachable volume of the manipulators.
    }
    \label{fig:scenarios}
    \vspace{-1em}
\end{figure*}

\begin{figure*}
    \centering
    \includegraphics[width=1.0\linewidth]{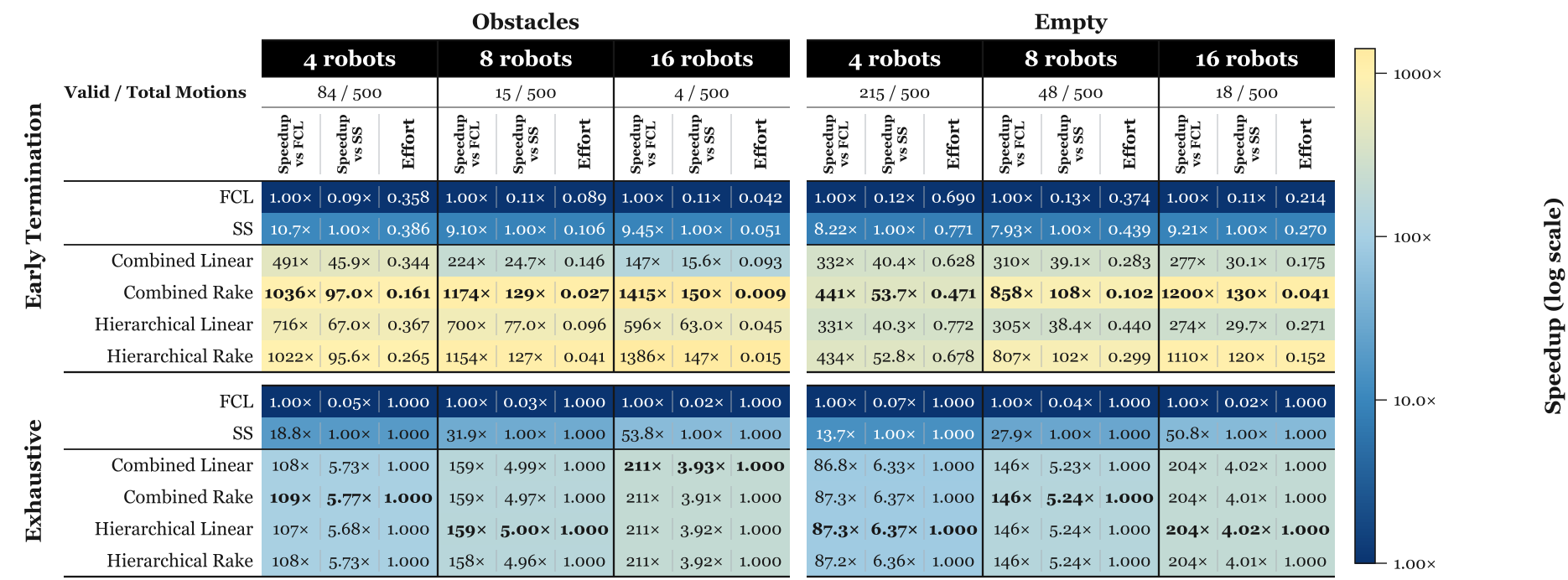}
    \caption{
    \small
    \textcolor{black}{Validation performance for 500 {\cageenv} motions with 4, 8, and 16 Pandas, with and without obstacles.
    The top row reports valid/total motions under the sphere geometry model (3\% of validaty-labels vary with {\fcl}).
    Early-termination trials stop at the first detected collision; 
    exhaustive trials execute all available validation checks.
    Each configuration reports speedup over {\fcl}, speedup over Sequential Sphere ({\seqsph}), and effort, the fraction of available checks executed. 
    Color encodes {\fcl}-relative speedup on a log scale.
    Bold marks the best value for each setting.
    }}
    \label{fig:validation-table}
    \vspace{-2em}
\end{figure*}

\subsection{\textcolor{black}{Experimental Design}}
\label{sec:experimental-design}

The experiments are designed to evaluate the three claims made in this paper: (i) vector-accelerated multi-robot primitives improve validation time, (ii) they improve {\mrmp} algorithm runtimes, and (iii) {\mv} and {\fc} should be distinct primitives.

\subsubsection{\textcolor{black}{Experiments}}

\textcolor{black}{
To test (i) and (iii), we sample 500 random motions with valid endpoints for three team sizes in the {\cageenv} scenario, both with and without obstacles.
We compare combined and hierarchical collision ordering with linear and rake packing under normal early termination and exhaustive validation, which evaluates every collision relation at every intermediate state.
We report valid/total motions, speedup over validation baselines, and the fraction of {\cc}s executed before termination.
}

\textcolor{black}{
To test (ii), we evaluate every planner in Table~\ref{tab:algs} with the vector-accelerated and baseline primitives.
The table in Fig.~\ref{fig:planner-table} reports successes $k/N$, median successful runtime, portion of successful-run time spent in {\mv} and {\fc}, and paired geometric-mean speedups in total and primitive time relative to each baseline.
Speedups use matched task-seed trials solved by both validation types; primitive timing includes the corresponding baseline implementations. 
Fig.~\ref{fig:planner-plots} shows all-trial runtime CDFs for one common team size per scenario; each curve’s terminal height gives its success rate, and a 1s reference marks subsecond solutions.
}

\subsubsection{\textcolor{black}{Scenarios}}
\label{sec:exp-scenarios}
\textcolor{black}{
We test increasing team sizes in the {\crossenv}, {\cageenv}, and {\heteroenv} scenarios to vary coordination and robot complexity (Fig.~\ref{fig:scenarios}).
{\cageenv} obstacles are cylinders; the mobile robots are spheres constrained to $z=0$.
{\crossenv} trials are given 10 seconds and {\cageenv} and {\heteroenv} trials are given 150 seconds to find a solution.
{\cageenv} uses five random tasks with 10 seeds each,
{\crossenv} uses 30 seeds, and {\heteroenv} uses 10 seeds.
}

\subsubsection{\textcolor{black}{Implementations and Baselines}}
\label{sec:exp-implementation}

\textcolor{black}{
In all of our experiments, we use three validation methods: {\fcl}~\cite{pan2012fcl}, Sequential Sphere ({\seqsph}), and {\vamrmp}.
{\fcl} uses the standard Panda meshes distributed with {\vamp}~\cite{thomason2024motions}, while {\vamrmp} and {\seqsph} both use the same {\vamp} sphere model.
{\fcl} and {\seqsph} scan states linearly and terminate on the first detected collision.
Thus, {\fcl}-{\seqsph} comparisons study the impact of the alternative geometry, whereas {\fcl}-{\vamrmp} comparisons consider all implementation changes.
Planner parameters are tuned for each scenario and are constant for all primitive implementations.
Priority orders are sampled randomly for PP-ST-RRT for each seed;
orders for {\heteroenv} ensure Pandas are planned for before flying spheres otherwise the spheres pass through the bases.
}

\textcolor{black}{All primitive and planner implementations} are written in C++ and built on the recently released {\ompl} 2.0~\cite{sucan2012the-open-motion-planning-library} with native {\vamp} integrations.
\textcolor{black}{
Each robot is limited to 1/128 units of Euclidean distance in its {\cspace} per timestep.
A composite motion uses the longest duration required by any robot, with all other robot motions time-scaled to start and finish simultaneously.
{\vamrmp} uses a fixed {\simd} vector width of $W=8$, so each batch contains up to eight configurations evaluated in parallel; 
$W$ is fixed by the implementation and was not tuned.}
\textit{The implementations will be made open-source upon acceptance of this paper to maintain a double-blind review.}
All experiments were conducted on a single Linux workstation (x86\_64, kernel 6.8) equipped with an Intel Core i7-14700F CPU (up to 5.4 GHz) and 32 GB of RAM.

\subsection{\textcolor{black}{Validation Experiments}}
\label{sec:validation-timing}

\textcolor{black}{
The table in Fig.~\ref{fig:validation-table} reports the relative speedup of {\vamrmp} and {\seqsph} over {\fcl} on the 500 randomly sampled motions (with valid endpoints).
When enabling early termination, the {\vamrmp} variants report 147$\times$-1415$\times$ speedup in validation time.
Among the {\vamrmp} variants, the rake packing strategy significantly outperforms the linear strategy, especially as the number of robots increases.
As the linear strategy is equivalent to adding robot-obstacle {\cc}ing to the {\fc} primitive, this justifies our claim that {\mv} and {\fc} should be distinct.
}

\textcolor{black}{
The presence of obstacles increased the relative performance of the hierarchical strategies.
While neither surpassed the combined rake strategy, this does suggest that the hierarchical rake strategy may be favorable in environments with greater obstacle density or fewer robot-robot collisions.
Even in the empty setting, only 18/500 of the 16-robot motions were valid due to self or robot-robot collisions.
}

\textcolor{black}{
Exhaustive validation disables early termination and checks every collision relation at every timestep, exposing full-scan implementation throughput independent of when a collision is encountered.
Here, disabling early termination exposes difference in geometry, validation implementation, and vectorized throughput, and there is minimal difference between {\vamrmp} variants.
For the rake variants, speedup increases with robot count under both early-terminating and exhaustive validation.
}

\textcolor{black}{
We use the combined rake strategy in all of the planner configurations in the following subsection.
}

\subsection{\textcolor{black}{Planner Experiments}}
\label{sec:planner-exp}

\begin{figure*}
\vspace{-2em}
    \centering
    \includegraphics[width=1.0\linewidth]{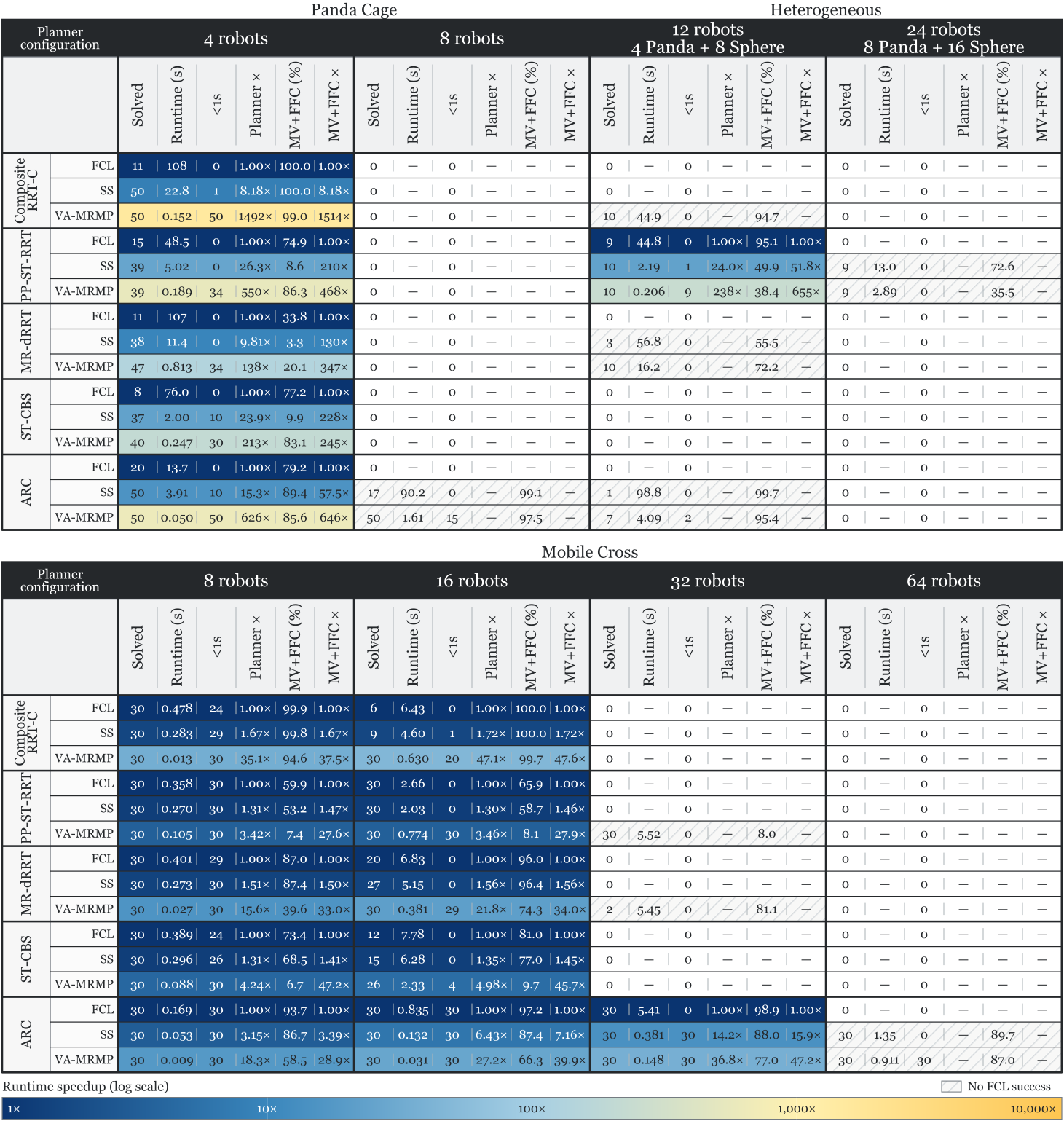}
    \caption{
    \small
    \textcolor{black}{
    Planner performance across all scenarios, methods, and validation implementations.
    Solved is the number of trials completed before timeout ({\cageenv}: N=50; {\heteroenv}: N=10; {\crossenv}: N=30); 
    $<1$s is the number of trials completed within one second.
    Runtime is the median time to first solution over successful trials, and MV+FFC (\%) is the median successful-run fraction spent in {\mv} and {\fc}.
    Speedups are paired geometric means of per-seed {\fcl} to {\seqsph} or {\vamrmp} ratios; 
    every seed solved by {\fcl} was also solved by {\seqsph} and {\vamrmp}, and every seed solved by {\seqsph} was also solved by {\vamrmp}. 
    Thus, no {\fcl}-successful seed is excluded from the {\fcl}-relative speedups. 
    Row color encodes planner speedup over {\fcl} on a log scale.
    Hatching marks backend successes with no paired {\fcl} success; 
    a dash indicates no successful trials or paired comparison.
    }}
    \label{fig:planner-table}
\vspace{-2em}
\end{figure*}

\begin{figure*}
    \centering
    \includegraphics[width=1.0\linewidth]{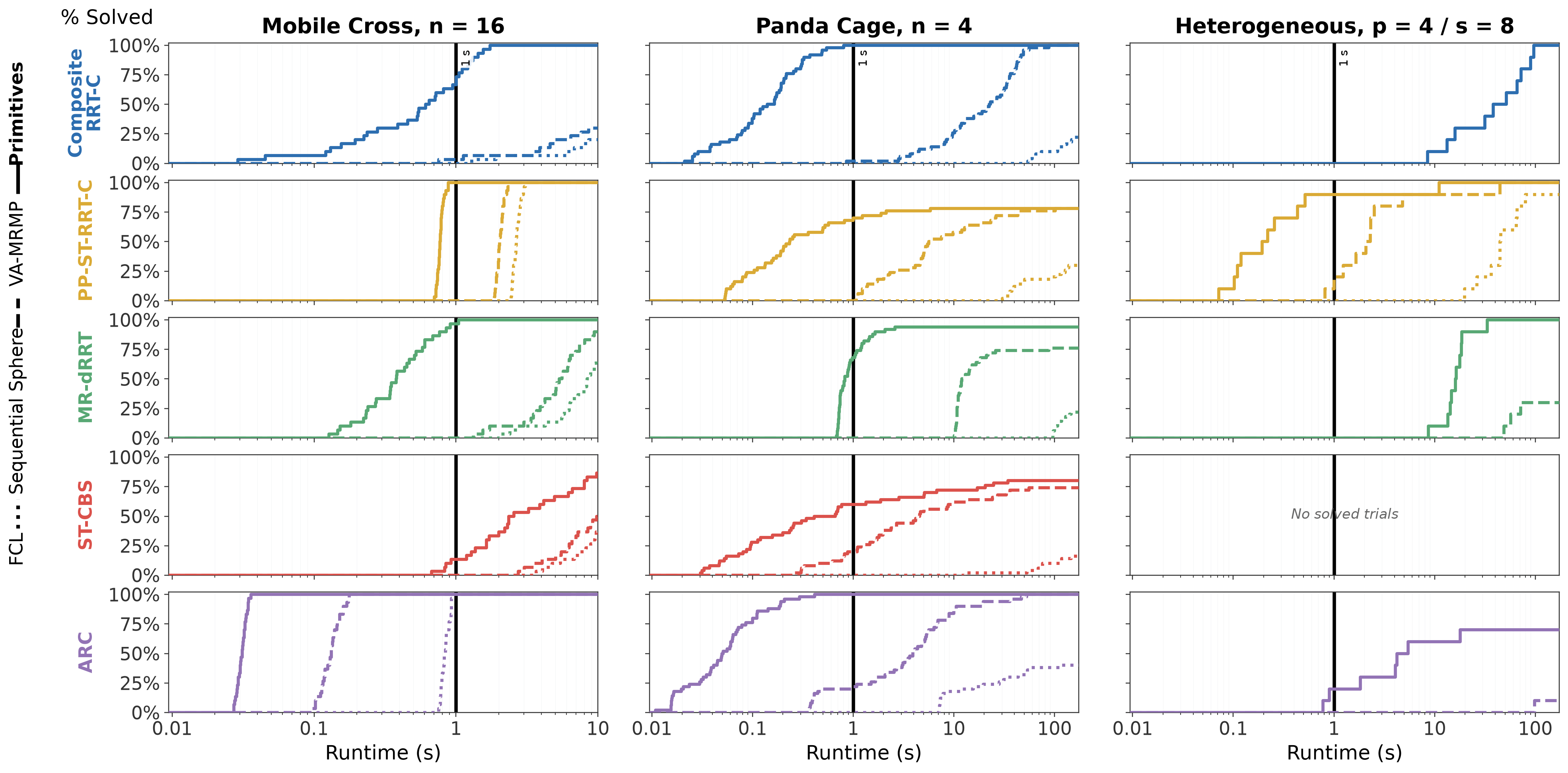}
    \caption{
    \small
    \textcolor{black}{
    All-trial cumulative distributions of runtimes for one common team size per scenario: 16 Mobile robots (N=30), 4 Pandas (N=50), and 4 Pandas with 8 spheres (N=10). 
    Rows correspond to planners; solid, dashed, and dotted curves denote {\vamrmp}, {\seqsph}, and {\fcl}, respectively.
    Curves use all trials as the denominator, so their terminal heights give success rates. 
    The vertical line marks one second; runtime is shown on a logarithmic scale.
    }}
    \label{fig:planner-plots}
    \vspace{-2em}
\end{figure*}

\textcolor{black}{
The comprehensive set of results for the planner experiments is provided in Fig.~\ref{fig:planner-table} with an illustrative plot for one team-size for each scenario in Fig.~\ref{fig:planner-plots}.
Across configurations with successful paired-trials, {\vamrmp} generally reduces median successful runtime and often increases the number of trials solved before timeout.
}

\textcolor{black}{
The speedups are larger in the more complex {\cageenv} scenario compared to the {\crossenv} scenario.
However, the {\crossenv} scenario continues the trend seen in Section~\ref{sec:validation-timing} where the speedup increases as the number of robots increases.
This corresponds with the increases portion of runtime spent on {\mv}/{\fc} as the cost of both primitives scales quadratically with the number of robots.
}

\textcolor{black}{
Composite RRT-C consistently demonstrates the greatest speedup in runtime.
This is due to Composite RRT-C runtime being almost entirely {\mv} operations as can be seen in the 4-robot {\cageenv} setting where it achieved 1492$\times$ speedup with 99\% of the runtime spent in {\mv}.
This also corresponds with the speedup seen in Fig.~\ref{fig:validation-table} for the rake strategies.
Meanwhile, ST-CBS, a method using the {\fc} primitive, has primitive and overall speedups closer to the linear strategies in Fig.~\ref{fig:validation-table}.
ARC, a method with both, reports speedups in-between.
}

\textcolor{black}{
PP-ST-RRT is the only method which demonstrates greater runtime speedup than {\mv}/{\fc} speedup.
This additional acceleration is from the single-robot {\vamp} operations.
These are also present in the roadmap generation of MR-dRRT, the UST-RRT queries of ST-CBS, and the initial individual motion plans in ARC and are not included in the reported {\mv}/{\fc} results.
}

\textcolor{black}{
In addition to the relative speedup improvement, {\vamrmp} moves each of the planning algorithms into the millisecond regime for nontrivial {\mrmp} instances (Fig.~\ref{fig:planner-plots}).
In the 4-robot {\cageenv} scenario, all five planners achieve subsecond median runtimes with over half their trials falling in the millisecond regime.
The same is true for four of the algorithms for the 16-robot mobile cross.
In the 8-robot {\cageenv} scenario, ARC has 15/50 trials finish in less than 1 second, and achieved a 4.29s median runtime on the 16-robot configuration over 48/50 successful trials.
In the {\heteroenv}, PP-ST-RRT solved 9/10 trials in less than 1 second and has a median runtime of 2.89s when the team size doubled.
}
\vspace{-1em}

\section{Conclusion}
\label{sec:conclusion}

\textcolor{black}{
In this paper, we have demonstrated that how you apply vector-acceleration to multi-robot validation matters, developed several strategies for doing so, and evaluated the relative speedup they offer across a representative set of algorithms on varied problem settings.
\textcolor{black}{
The vector-accelerated validation developed here brings several multi-robot planning algorithms and problem instances into the millisecond regime.
}
Open-source code of the vector-accelerated primitives and {\mrmp} algorithms will be made available upon acceptance.
}
\vspace{-2em}

\bibliographystyle{IEEEtran}
\bibliography{robotics.bib}

@string{TRA="{IEEE} Trans.\ Robot.\ Automat."}

@string{CACM="Communications of the ACM"}

@string{ICRA="Proc.\ {IEEE} Int.\ Conf.\ Robot.\ Autom.\ ({ICRA})"}

@string{IROS="Proc.\ {IEEE} Int.\ Conf.\ Intel.\ Rob.\ Syst.\ ({IROS})"}

@book{c-crmp-88
  , author = "J. F. Canny"
    , title = "The Complexity of Robot Motion Planning"
    , publisher = "MIT Press"
    , address = "Cambridge, MA"
    , year = 1988
}

@article{kslo-prpp-96,
    author = "L. E. Kavraki and P. \v{S}vestka and J. C. Latombe and M. H. Overmars",
    title = "Probabilistic Roadmaps for Path Planning in High-Dimensional Configuration Spaces",
    journal = TRA,
    volume = 12,
    number = 4,
    year = 1996,
    month = aug,
    pages = "566-580"
}

@article {lw-apcfpapo-79
  , author   =  "T. {Lozano-P\'{e}rez} and M. A. Wesley"
    , title =  "An Algorithm for Planning Collision-Free Paths
    Among Polyhedral Obstacles"
    , journal  = CACM
    , volume   = "22"
    , number = "10"
    , pages =  "560--570"
    , month = oct
    , year  =  1979
}

@INPROCEEDINGS{sl-uppccdpmrs-2002,
  author={Sanchez, G. and Latombe, J.-C.},
  booktitle=ICRA,
  title={Using a PRM planner to compare centralized and decoupled planning for multi-robot systems},
  year={2002},
  volume={2},
  pages={2112--2119}
}

@inproceedings{bk-ppulp-00
, author = {Robert Bohlin and Lydia E. Kavraki }
, title = {Path Planning Using Lazy PRM}
, booktitle = ICRA
, month = apr
, year = {2000}
}

@article{ssfs-cbsfomap-15,
  title={Conflict-based search for optimal multi-agent pathfinding},
  author={Sharon, Guni and Stern, Roni and Felner, Ariel and Sturtevant, Nathan R},
  journal={Artificial Intelligence},
  volume={219},
  pages={40--66},
  year={2015},
  publisher={Elsevier}
}

@article{smsa-rmmpucs-21,
  title={Representation-optimal multi-robot motion planning using conflict-based search},
  author={Solis, Irving and Motes, James and Sandstr{\"o}m, Read and Amato, Nancy M},
  journal={IEEE Robotics and Automation Letters},
  volume={6},
  number={3},
  pages={4608--4615},
  year={2021},
  publisher={IEEE}
}

@article{ss-otpmpiigtfctporam-83,
  title={On the “piano movers” problem. II. General techniques for computing topological properties of real algebraic manifolds},
  author={Schwartz, Jacob T and Sharir, Micha},
  journal={Advances in applied Mathematics},
  volume={4},
  number={3},
  pages={298--351},
  year={1983},
  publisher={Elsevier}
}

@article{ssh-faniaehdrfeoirimm-16,
  title={Finding a needle in an exponential haystack: Discrete RRT for exploration of implicit roadmaps in multi-robot motion planning},
  author={Solovey, Kiril and Salzman, Oren and Halperin, Dan},
  journal={The International Journal of Robotics Research},
  volume={35},
  number={5},
  pages={501--513},
  year={2016},
  publisher={SAGE Publications Sage UK: London, England}
}

@inproceedings{thomason2024motions,
  title={Motions in microseconds via vectorized sampling-based planning},
  author={Thomason, Wil and Kingston, Zachary and Kavraki, Lydia E},
  booktitle={2024 IEEE international conference on robotics and automation (ICRA)},
  pages={8749--8756},
  year={2024},
  organization={IEEE}
}

@inproceedings{sim2025st,
  title={ST-CBS: Spatio-Temporal Conflict Based Search in Continuous Space for Multi-Agent Pathfinding},
  author={Sim, Joonyeol and Lim, Seonghyeon and Nam, Changjoo},
  booktitle={Proceedings of the 40th ACM/SIGAPP Symposium on Applied Computing},
  pages={815--822},
  year={2025}
}

@inproceedings{grothe2022st,
  title={St-rrt*: Asymptotically-optimal bidirectional motion planning through space-time},
  author={Grothe, Francesco and Hartmann, Valentin N and Orthey, Andreas and Toussaint, Marc},
  booktitle={2022 International Conference on Robotics and Automation (ICRA)},
  pages={3314--3320},
  year={2022},
  organization={IEEE}
}

@inproceedings{kuffner2000rrt,
  title={RRT-connect: An efficient approach to single-query path planning},
  author={Kuffner, James J and LaValle, Steven M},
  booktitle={Proceedings 2000 ICRA. Millennium conference. IEEE international conference on robotics and automation. Symposia proceedings (Cat. No. 00CH37065)},
  volume={2},
  pages={995--1001},
  year={2000},
  organization={IEEE}
}

@article{solis2024adaptive,
  title={Adaptive robot coordination: A subproblem-based approach for hybrid multi-robot motion planning},
  author={Solis, Irving and Motes, James and Qin, Mike and Morales, Marco and Amato, Nancy M},
  journal={IEEE Robotics and Automation Letters},
  volume={9},
  number={8},
  pages={7238--7245},
  year={2024},
  publisher={IEEE}
}

@inproceedings{kerimov2025si,
  title={SI-RRT and ST-RRT* for Prioritized Multi-Manipulator Planning: Empirical Evaluation},
  author={Kerimov, Nuraddin and Onegin, Aleksandr and Yakovlev, Konstantin},
  booktitle={International Conference on Interactive Collaborative Robotics},
  pages={371--384},
  year={2025},
  organization={Springer}
}

@article{sucan2012the-open-motion-planning-library,
    Author = {Ioan A. {\c{S}}ucan and Mark Moll and Lydia E. Kavraki},
    Doi = {10.1109/MRA.2012.2205651},
    Journal = {{IEEE} Robotics \& Automation Magazine},
    Month = {December},
    Number = {4},
    Pages = {72--82},
    Title = {The {O}pen {M}otion {P}lanning {L}ibrary},
    Note = {\url{https://ompl.kavrakilab.org}},
    Volume = {19},
    Year = {2012}
}

@inproceedings{sanchez2003single,
  title={A single-query bi-directional probabilistic roadmap planner with lazy collision checking},
  author={S{\'a}nchez, Gildardo and Latombe, Jean-Claude},
  booktitle={Robotics Research: The Tenth International Symposium},
  pages={403--417},
  year={2003},
  organization={Springer}
}

@inproceedings{pan2012fcl,
  title={FCL: A general purpose library for collision and proximity queries},
  author={Pan, Jia and Chitta, Sachin and Manocha, Dinesh},
  booktitle={2012 IEEE international conference on robotics and automation},
  pages={3859--3866},
  year={2012},
  organization={IEEE}
}

@article{huang2025prrtc,
  title={prrtc: Gpu-parallel rrt-connect for fast, consistent, and low-cost motion planning},
  author={Huang, Chih H and Jadhav, Pranav and Plancher, Brian and Kingston, Zachary},
  journal={arXiv preprint arXiv:2503.06757},
  year={2025}
}

@article{sundaralingam2023curobo,
  title={curobo: Parallelized collision-free minimum-jerk robot motion generation},
  author={Sundaralingam, Balakumar and Hari, Siva Kumar Sastry and Fishman, Adam and Garrett, Caelan and Van Wyk, Karl and Blukis, Valts and Millane, Alexander and Oleynikova, Helen and Handa, Ankur and Ramos, Fabio and others},
  journal={arXiv preprint arXiv:2310.17274},
  year={2023}
}

@inproceedings{huang2026vampmr,
  title        = {VAMP-MR: Vector-Accelerated Motion Planning and Execution for Multi-Robot-Arms},
  author       = {Huang, Philip and Gao, Chenrui and Li, Jiaoyang},
  booktitle    = {IEEE/RSJ International Conference on Intelligent Robots and Systems (IROS)},
  year         = {2026}
}

\end{document}